\pdfoutput = 1
\documentclass{article}

\usepackage[final,nonatbib]{nips_2016}

\usepackage[utf8]{inputenc} 
\usepackage[T1]{fontenc}    
\usepackage{hyperref}       
\usepackage{url}            
\usepackage{booktabs}       
\usepackage{amsfonts}       
\usepackage{nicefrac}       
\usepackage{microtype}      

\usepackage{amsmath,amssymb} 
\usepackage{mathtools}
\usepackage{subcaption}

\usepackage{comment}
\usepackage{multirow}
\usepackage{multicol}
\usepackage{bbm}
\usepackage{color}

\usepackage{algorithm}
\usepackage{algorithmic}

\title{Scalable Gaussian Processes for Supervised Hashing}

\author{
  Bahadir Ozdemir \\
  Department of Computer Science\\
  University of Maryland \\
  College Park, MD 20742 \\
  \texttt{ozdemir@cs.umd.edu} \\
  \And
  Larry S. Davis \\
  Institute for Advanced Computer Studies\\
  University of Maryland \\
  College Park, MD 20742 \\
  \texttt{lsd@umiacs.umd.edu} \\
}

\DeclareMathOperator*{\argmax}{\arg\!\:\max}

\begin{document}

\maketitle

\begin{abstract}
We propose a flexible procedure for large-scale image search by hash functions with kernels. Our method treats binary codes and pairwise semantic similarity as latent and observed variables, respectively, in a probabilistic model based on Gaussian processes for binary classification. We present an efficient inference algorithm with the sparse pseudo-input Gaussian process (SPGP) model and parallelization. Experiments on three large-scale image dataset demonstrate the effectiveness of the proposed hashing method, Gaussian Process Hashing (GPH), for short binary codes and the datasets without predefined classes in comparison to the state-of-the-art supervised hashing methods.
\end{abstract}

\section{Introduction}
\label{sec:introduction}

As the size of online image datasets gets larger, the problem of image retrieval has attracted more attention from researchers in computer vision, machine learning, and information retrieval. Considering an image dataset with millions of images, image retrieval becomes a seemingly intractable problem for exhaustive similarity search algorithms due to their linear time complexity. Although $k$-d trees are useful data structures for low-dimensional nearest neighbor search, they are not as applicable to high-dimensional image descriptors. Hashing methods, which encode high-dimensional descriptors into compact binary strings, are attractive alternatives due to their sublinear time search complexity and efficient storage capacity. Methods for learning binary codes can be roughly grouped into two categories: Unsupervised hashing where binary codes are designed to preserve similarity in the original feature space; and supervised hashing where binary codes are constructed to capture semantic similarity between images.

Unsupervised hashing techniques can be categorized as data independent and dependent schemes. Locality Sensitive Hashing (LSH) \cite{Gionis:1999wf} is the most widely known data-independent hashing technique, which has also been extended to hashing with kernels (KLSH) \cite{Kulis:2012ey}. Notable data-dependent hashing methods are Spectral Hashing (SH) \cite{Weiss:2008tu}, Principal Component Analysis based Hashing (PCAH) \cite{Wang:2010jk} and Iterative Quantization (ITQ) \cite{Gong:2013kp}. Unfortunately, unsupervised hashing methods unavoidably suffer from the semantic gap, which is the difference between low-level image descriptors and high-level semantic representation of an image, when they are employed for image retrieval. Therefore, most recent unsupervised methods \cite{Nitish:2012dl, Rastegari:2013ue, Ozdemir:2014wi} focus on multimodal data to bridge this gap by utilizing both textual and visual features. Yildirim and Jacobs \cite{Yildirim2012-YILARA} propose an integrative model based on Indian Buffet Process (IBP) to connect features in different modalities by common abstract features. In our previous work \cite{Ozdemir:2014wi}, we utilize this model for multimodal retrieval.

Supervised Hashing methods usually rely on pairwise labels, where $+1$ and $-1$ indicate that two points are similar and dissimilar, respectively. Furthermore, some methods use 0 for points which are neighbors in the metric space without being semantically similar. Supervised Hashing with Kernels (KSH) \cite{Liu:2012ii} tries to minimize the Hamming distances between similar pairs and simultaneously maximize them for dissimilar pairs using inner products of binary codes. FastHash \cite{Lin:2015gs} employs decision trees as hash functions and a GraphCut based method for the same type of optimization problem. Minimal loss hashing (MLH) \cite{Norouzi:2011mf} uses a structured SVM framework to generate binary codes. Supervised Hashing with latent factor models (LFH) \cite{Zhang:2014jr} uses latent factors for learning binary codes in a probabilistic framework. Supervised discrete hashing (SDH) \cite{Shen:2015fs} learns binary codes based on linear classification. The main problem of most supervised hashing methods is overfitting to training data. Despite different formulations, many supervised hashing methods in practice find one binary string pattern for each class and the same binary hash code is assigned to all images in one class.

The Gaussian process is an elegant model based on Bayesian nonparametrics for nonlinear regression and classification. Similar to SVM, which is employed in \cite{Rastegari:2013ue, Norouzi:2011mf}, binary Gaussian process classification (GPC) model \cite{Kuss:2005wx} can be utilized for hashing in a Bayesian framework. In this paper, we propose a supervised hashing method with kernels based on the GPC model, so we call the proposed approach \emph{Gaussian Process Hashing} (GPH). Unlike LFH, which first optimizes latent factors independently from image features and then learns a weight matrix to map these features to the latent factors, we use a fully probabilistic approach for learning binary codes from data. In the GPH model, the variables corresponding to the binary classes in the original GPC model become latent variables and a new level of variables that correspond to the pairwise similarity between data points is integrated into the model. Averaging over nonlinear classification models by Gaussian processes is our motivation behind the GPH model to overcome the overfitting problem of supervised hashing. Our contributions can be summarized as follows:
\begin{itemize}
\item We propose a flexible Bayesian nonparametric model based on binary Gaussian process classification for supervised learning of binary codes with better generalization.

\item We utilize the GPC predictive distribution to define a hash function by means of hyperplanes with kernels.

\item We developed a scalable parallel inference algorithm for the proposed model using a sparse approximation to the GPC model via a hybrid approach combining MCMC and message passing.
\end{itemize}

The rest of the paper is organized as follows: Section \ref{sec:method} describes the GPH. The inference algorithm for the method is explained in Section \ref{sec:inference}. Section \ref{sec:experiments} demonstrates performance evaluation and also the comparison of our method with the state-of-the-art supervised hashing methods, and Section \ref{sec:conclusion} provides conclusions regarding the proposed method.

\section{Gaussian Process Hashing}
\label{sec:method}

\subsection{Problem Definition}

Given a set of data points $\mathcal{X}= \{\mathbf{x}_1,\ldots,\mathbf{x}_n\}$, each $\mathbf{x}_i\in\mathbb{R}^d$, hashing aims to find a function $H$ that maps data points from $\mathbb{R}^d$ to $m$-dimensional Hamming space with respect to some optimization criteria. Let $\mathcal{S}=\{s_{il}\}$ denote the set of similarity labels for pairs of data points where $s_{il}=+1$ if $\mathbf{x}_i$ and $\mathbf{x}_l$ are similar and $s_{il}=-1$ otherwise, then a desirable hash function yields a smaller Hamming distance between $H(\mathbf{x}_i)$ and $H(\mathbf{x}_l)$ when $s_{il}=+1$ and a larger one when $s_{il}=-1$. Many techniques \cite{Gionis:1999wf, Gong:2013kp, Liu:2012ii, Kulis:2012ey} construct such a hash function $H:\mathbb{R}^d \rightarrow\{-1,+1\}^m$ by combining $m$ independent binary encoding functions $h_1(\mathbf{x}),\ldots, h_m(\mathbf{x})$ such that $h_j(\mathbf{x})=\mathrm{sgn}(\mathbf{w}_j^\top \mathbf{x}+b_j)$ where $\mathbf{w}_j\in\mathbb{R}^d$ is a hyperplane, $b_j\in\mathbb{R}$ is an intercept for $j=1,\ldots,m$ and
\begin{equation*}
    \mathrm{sgn}(v)=
    \begin{cases}
        +1 & \textrm{if $v\geq0$},\\
        -1 & \textrm{otherwise.}
    \end{cases}
\end{equation*}
Some hashing methods \cite{Kulis:2012ey, Liu:2012ii} employ a kernel function $k:\mathbb{R}^d\times\mathbb{R}^d\rightarrow\mathbb{R}$ such that each binary encoding hash function has the form $h_j(\mathbf{x})=\mathrm{sgn}(\mathbf{w}_j^\top\mathbf{k}+b_j)$ where $\mathbf{k}=[k(\mathbf{x},\bar{\mathbf{x}}_1),\ldots,k(\mathbf{x},\bar{\mathbf{x}}_r)]^\top$ and $\{\bar{\mathbf{x}}_1,\ldots,\bar{\mathbf{x}}_r\}\subseteq\mathcal{X}$.

\subsection{A Probabilistic Approach}

The hash function $H$ can be modeled as a latent function that builds a bridge between $\mathcal{X}$ and $\mathcal{S}$, \emph{i.e.} the observed data $\mathcal{D} = \{\mathcal{X},\mathcal{S}\}$. Let $\mathbf{Y}$ be a random binary matrix of size $n\times m$ where $Y_{ij}=h_j(\mathbf{x}_i)$ and $\mathbf{y}_j$ is the $j$th column of $\mathbf{Y}$, \emph{i.e.} shorthand for the values of the hash function. Then using Bayes' rule we can write
\begin{equation}
\label{eq:pY_D}
    p(\mathbf{Y}|\mathcal{D}) = \frac{p(\mathcal{S}|\mathbf{Y})\:\! p(\mathbf{Y}|\mathcal{X})}
    {p(\mathcal{S}|\mathcal{X})}.
\end{equation}
Given the binary codes $\mathbf{Y}$, the similarity labels $\mathcal{S}$ are assumed to be independent Bernoulli variables. Let $\mathbf{V}=\mathbf{Y}\mathbf{Y}^\top$, \emph{i.e.} $V_{il}$ is the inner product of binary codes for sample $i$ and sample $l$; then the joint likelihood factorizes as
\begin{equation}
\label{eq:pS_Y}
    p(\mathcal{S}|\mathbf{Y})=\prod_{(i,l)}    p(s_{il}\vert V_{il}) = 
    \prod_{i=2}^{n-1}\,\prod_{l=i+1}^n p(s_{il}\vert V_{il}).
\end{equation}
We use the probit model to define $p(s_{il}=+1\:\!\vert\:\! V_{il})=\Phi(\sigma_y  V_{il})$ where $\Phi$ denotes the cumulative distribution function of the standard normal distribution and $\sigma_y>0$ is a scaling parameter. The individual likelihood terms can be written as $p(s_{il}\vert V_{il}) = \Phi(\sigma_y s_{il}\:\!V_{il})$ due to the symmetry of $\Phi$ around zero. Note that $p(\mathcal{S}|\mathbf{Y})$ is maximized when the columns of $\mathbf{Y}$ are orthogonal and images of the same semantic class have the same binary embeddings. The prior term of \eqref{eq:pY_D} can be designed as a Gaussian process model for binary classification as described in the following section.

\subsection{The Gaussian Process Model for Binary Classification}

In this section we briefly describe Gaussian Process Classification (GPC); for more details see \cite{Kuss:2005wx, Rasmussen:2006vz, Nickisch:2008vx}. The GPC model is a discriminative Bayesian classifier that models $p(y|\mathbf{x})$ as a Bernoulli distribution for a given data point $\mathbf{x}$. The class membership probability is characterized by an underlying latent function $f(\mathbf{x})$. The value of the latent function is mapped into the unit interval by a sigmoid function $\sigma:\mathbb{R}\rightarrow[0,1]$ such that the probability $p(y=+1\,|\,\mathbf{x})$ becomes $\sigma(f(\mathbf{x}))$. We again prefer the probit model $p(y=+1\,|\,\mathbf{x})=\Phi(f(\mathbf{x}))$ due to analytical convenience of the inference algorithm.

In GPH, there exist $m$ latent functions, one for each bit $j$ in $H$, which are assumed to be a priori independent. Let $\mathbf{F}$ be a random real-valued matrix of size $n\times m$ where $F_{ij}= f_j(\mathbf{x}_i)$ and $\mathbf{f}_j$ be the $j$th column of $\mathbf{F}$ \emph{i.e.} shorthand for the values of the latent functions; then the binary codes $\mathbf{Y}$ are independent Bernoulli variables conditioned on $\mathbf{F}$, so the joint likelihood factorizes as
\begin{equation}
\label{eq:pY_F}
    p(\mathbf{Y}|\mathbf{F})     = \prod_{j=1}^m p(\mathbf{y}_j|\mathbf{f}_j) = 
    \prod_{j=1}^m \prod_{i=1}^n \Phi(Y_{ij}\:\!F_{ij})
\end{equation}

We place a zero mean Gaussian process prior on each latent function $f_j$ to obtain $p(y_j=+1\,\vert\,\mathbf{x})=\frac{1}{2}$ for $j=1,\ldots,m$ \cite{Kuss:2005wx}. Note that the probability for each similarity label $p(s_{il}=+1\,\vert\,\mathbf{x}_i,\mathbf{x}_l)$ eventually becomes $\frac{1}{2}$ as well.

Through this stochastic process, each data point $\mathbf{x}_i$ is associated with $m$ random variables $\{f_j(\mathbf{x}_i)\}_{j=1}^m$. Considering the entire dataset $\mathcal{X}$, we therefore have $m$ independent multivariate Gaussian distributions. The joint distribution of latent function values for the $j$th bit is $p(\mathbf{f}_j\vert\mathcal{X})=\mathcal{N}(\mathbf{f}_j\vert\mathbf{0},\mathbf{K})$ where the covariance matrix is constructed from a kernel function $K_{il}=k(\mathbf{x}_i,\mathbf{x}_l)$ that depends on a set of kernel hyperparameters $\boldsymbol\theta$. In our experiments, we use the squared exponential covariance function with the isotropic distance measure of the form:
\begin{equation*}
    k(\mathbf{x},\mathbf{x}') = \sigma_f^2\:\!\exp\bigl(-\tfrac{1}{2\ell^2}\|\mathbf{x}-\mathbf{x}'\|^2\bigr)
\end{equation*}
such that $\boldsymbol\theta=\{\sigma_f,\ell\}$ where $\sigma_f^2$ and $\ell$ are referred as the signal variance and the characteristic length-scale, respectively.

Given $\Theta=\{\sigma_y,\boldsymbol\theta\}$, we can compute the posterior distribution over the latent function values using the likelihood \eqref{eq:pY_F} and Bayes' rule as
\begin{equation}
\label{eq:pF_D}
\begin{split}
    p(\mathbf{F}|\mathcal{D}) &=
    \frac{p(\mathcal{S}|\mathbf{F})\:\!p(\mathbf{F}|\mathcal{X})}{p(\mathcal{D})} = \frac{\sum_\mathbf{Y} p(\mathcal{S}|\mathbf{Y}) \:\! p(\mathbf{Y}|\mathbf{F}) \:\!
    p(\mathbf{F}|\mathcal{X})}{p(\mathcal{D})}.
\end{split}
\end{equation}
Unfortunately, neither the posterior $p(\mathbf{F}|\mathcal{D})$ nor the marginal likelihood $p(\mathcal{D})$ can be computed analytically. Therefore, we approximate the posterior $p(\mathbf{F}|\mathcal{D})$ by joint Gaussian distributions $q(\mathbf{F}) = \prod_{j=1}^m\mathcal{N}(\mathbf{f}_j|\mathbf{m}_j, \mathbf{A}_j)$. Note that the approximate posterior distribution for each bit is independent of other bits. The details of the Gaussian approximation are presented in Section~\ref{sec:inference}.

\subsection{Predictions for Queries}

The main purpose of hashing models is to predict binary codes for queries. Prediction at a query input $\mathbf{x}_*$ is made by marginalizing out over $\mathbf{F}$ in the joint distribution $p(\mathbf{F},\mathbf{F}_*|\mathcal{D},\mathbf{x}_*)$. This can be done separately for each bit $j$ similar to \cite{Kuss:2005wx} because of the independence in the approximate posterior distribution:
\begin{equation}
\begin{split}
    q(f_{*j}\:\!\vert\:\!\mathcal{D},\mathbf{x}_*) &=
    \int p(f_{*j}\:\!\vert\:\!\mathbf{f}_j,\mathcal{X},\mathbf{x}_*)\:\!
    q(\mathbf{f}_j|\mathcal{D})\,d\mathbf{f}_j = \mathcal{N}(f_{*j}\:\!\vert\:\!\mu_{*j},\sigma_{*j}^2)
\end{split}    
\end{equation}
with mean and variance:
\begin{subequations}
\begin{align}
    \label{eq:muj}
    \mu_{*j} &= \mathbf{k}_*^\top \mathbf{K}^{-1}\mathbf{m}_j \\
    \sigma_{*j}^2 &= k(\mathbf{x}_*,\mathbf{x}_*) - \mathbf{k}_*^\top(\mathbf{K}^{-1} - 
    \mathbf{K}^{-1}\mathbf{A}_j\mathbf{K}^{-1})\,\mathbf{k}_*
\end{align}
\end{subequations}
where $\mathbf{k}_* = [k(\mathbf{x}_1,\mathbf{x}_*),\ldots,k(\mathbf{x}_n,\mathbf{x}_*)]^\top$
is a vector of prior covariances between the query input $\mathbf{x}_*$ and training inputs $\mathcal{X}$. Finally, the approximate predictive distribution for each binary encoding is given as follows \cite{Rasmussen:2006vz}:
\begin{equation}
\begin{split}
    q(y_{*j}\:\!\vert\:\!\mathcal{D},\mathbf{x}_*) &= \int p(y_{*j}\:\!\vert\:\!f_{*j})\,
    q(f_{*j}\:\!\vert\:\!\mathcal{D},\mathbf{x}_*)\,df_{*j}\\
    &= \int \Phi(y_{*j}\:\!f_{*j})\:\!\mathcal{N}(f_{*j}\:\!\vert\:\!\mu_{*j},\sigma_{*j}^2)\,df_{*j}
    = \Phi\biggl(\frac{y_{*j}\:\!\mu_{*j}}{\sqrt{\smash[b]{1+\sigma_{*j}^2}}}\biggr).
\end{split}
\end{equation}
This probability is maximized for each binary encoding by
\begin{equation}
\hat{y}_{*j}=\argmax_{y_{*j}\in\{\pm1\}} \, q(y_{*j}\vert\mathcal{D},\mathbf{x}_*) = \mathrm{sgn}(\mu_{*j})
\end{equation}
As a result, we introduce our hash function in the form of binary encodings using \eqref{eq:muj}:
\begin{equation}
\label{eq:hash}
    h_j(\mathbf{x}_*) = \mathrm{sgn}(\mathbf{w}_j^\top \mathbf{k}_*)
\end{equation} 
where $\mathbf{w}_j=\mathbf{K}^{-1}\mathbf{m}_j$,~~for $j=1,\ldots,m$. Our has function consists of only multiplication between weight matrix $\mathbf{W}=[\mathbf{w}_1,\ldots,\mathbf{w}_m]$ and query kernel vector $\mathbf{k}_*$, and then sign operation. Consequently, our hash function does not contain the costly cumulative distribution function $\Phi$.

\subsection{Sparse Approximation}

\begin{figure*}[!ht]
\centering
\includegraphics[width=.95\linewidth]{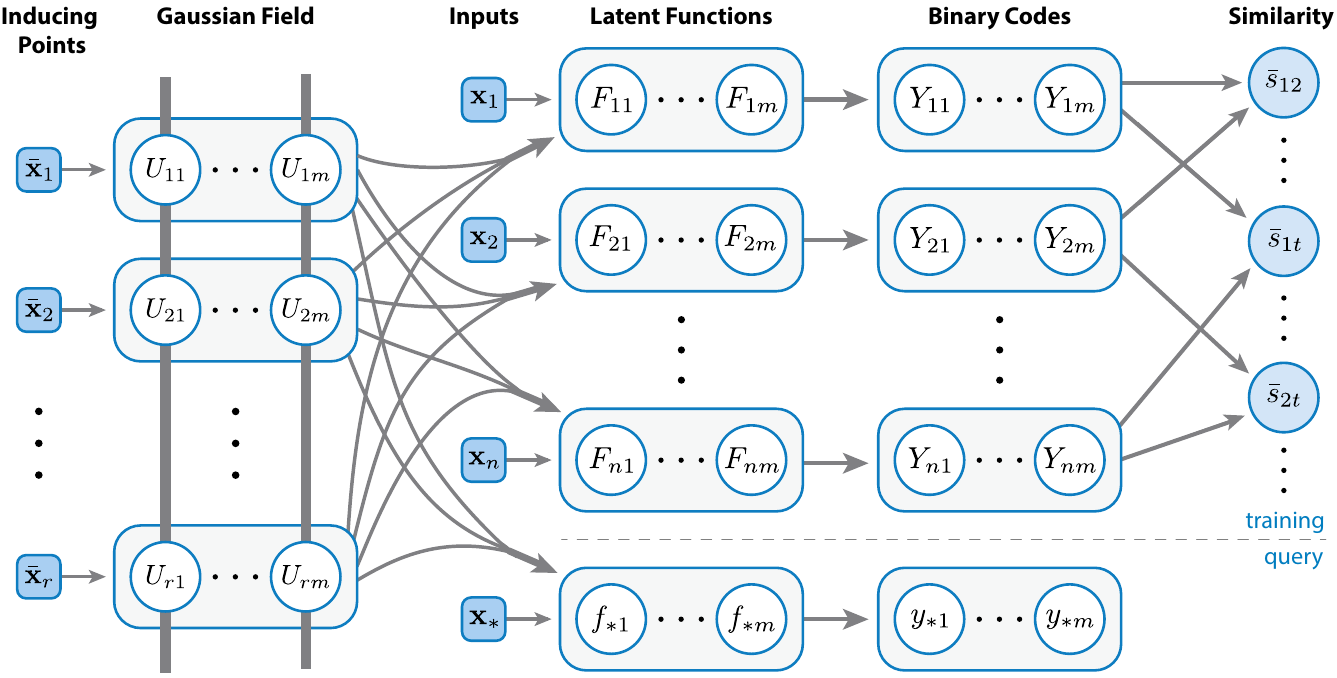}
\caption{Graphical model for the Gaussian Process Hashing where circles indicate random variables, shaded circles denote observed values. The thick vertical bars represent a set of fully connected nodes.}
\label{fig:graphical_model}
\end{figure*}

 To accelerate training and query times, we employ a sparse approximation to the full Gaussian process known as the fully independent training conditional (FITC) approximation \cite{Candela:2005wp, Snelson:2005wc}. Let $\bar{\mathcal{X}}=\{\bar{\mathbf{x}}_1,\ldots,\bar{\mathbf{x}}_r\}$, each $\bar{\mathbf{x}}_i\in\mathbb{R}^d$, which might be a subset of $\mathcal{X}$, and associated latent function values $\mathbf{U}$, analogous to $\mathbf{F}$. Then we obtain the FITC approximation for each bit $j$ as below: 
\begin{equation}
\begin{split}
    p(\mathbf{f}_j,f_{*j}\:\!\vert\:\!\mathcal{X},\bar{\mathcal{X}}, \mathbf{x}_*) &= 
    \int p(\mathbf{f}_j,f_{*j}\:\!\vert\:\!\mathbf{u}_j,\mathcal{X},\mathbf{x}_*)\:\!
    p(\mathbf{u}_j|\bar{\mathcal{X}})\,d\mathbf{u}_j\\ 
    &\approx \int  q(\mathbf{f}_j\:\!\vert\:\!\mathbf{u}_j,\mathcal{X})\:\!
    q(f_{*j}|\mathbf{u}_j,\mathbf{x}_*)\:\!p(\mathbf{u}_j|\bar{\mathcal{X}})\,d\mathbf{u}_j
\end{split}
\end{equation}
where $p(\mathbf{u}_j|\bar{\mathcal{X}})=\mathcal{N}(\mathbf{u}_j\:\!\vert\:\!\mathbf{0},\mathbf{K_{uu}})$. Marginalizing over the exact prior on $\mathbf{u}_j$ in the final expression yields
\begin{equation}
    q(\mathbf{f}_j|\mathcal{X})= \mathcal{N}\bigl(\mathbf{f}_j\:\!\vert\:\!\mathbf{0},
    \mathbf{Q_{ff}} + \mathsf{diag}(\mathbf{K_{ff}}-\mathbf{Q_{ff}})\bigr)
\end{equation}
where $\mathbf{Q_{ab}}=\mathbf{K_{au}}\mathbf{K}_\mathbf{uu}^{-1}\mathbf{K_{ub}}$ for $j=1,\ldots,m$. The computations of the FITC approximation for binary classification are explained thoroughly in \cite{NaishGuzman:2007uf}. We can define binary encodings similar to those of \eqref{eq:hash} for the FITC model as well. However, the hyperplanes $\mathbf{w}_j\in\mathbb{R}^r$ and $\mathbf{k}_*$ will be defined between the query input $\mathbf{x}_*$ and inducing inputs $\bar{\mathcal{X}}$ since training data and queries are conditionally independent given $\mathbf{U}$ in the FITC approximation. Therefore, the communication between them is established through the bottleneck of inducing inputs \cite{NaishGuzman:2007uf}. Similarly, we employ a sparse set in place of all pairwise similarities, $\mathcal{S}$, in \eqref{eq:pS_Y}. We randomly sample $t$ images, called \emph{representatives}, from the dataset and use the pairwise similarities only between those representatives and the entire training set. The sparse similarity set is denoted by $\mathcal{\bar{S}}$. By these sparsity changes, the time complexity of Gaussian process hashing is reduced from $\mathcal{O}(n^3m)$ to $\mathcal{O}\bigl(nm(r^2+t)\bigr)$. The graphical model of our hashing approach with sparse approximation is shown in Figure~\ref{fig:graphical_model}.

\section{Inference}
\label{sec:inference}

We follow the inference approach from \cite{DoshiVelez:2009tf}, which is a hybrid of message passing and Gibbs sampling. This technique has some advantages over a compound inference regarding convergence and efficiency. Our inference algorithm alternates between two phases: 
\begin{itemize}
    \item Updating site parameters of the approximate distribution by Expectation Propagation (EP) \cite{Minka:2001we} for each bit $j$ in parallel, and
    \item Sampling each entry of $\mathbf{Y}$ by Gibbs sampling.
\end{itemize}

The main idea behind the EP algorithm is to minimize the Kullback-Leibler (KL) divergence, which is achieved by matching of moments, at each step by adjusting site parameters. The details of EP are presented broadly by Gelman \emph{et al.} \cite{Gelman:2014ww}. In the first phase of our inference algorithm, the problem is reduced to Gaussian process classification since we know the values of the binary matrix $\mathbf{Y}$ thanks to the Gibbs sampler. Therefore, we adopt the new scalable inference scheme by Hernandez-Lobato and Hernandez-Lobato \cite{Hernandez:2016}, which is defined in a distributed and stochastic fashion, for the FITC model. We apply that scalable EP algorithm in a serial and stochastic setting for a full update of site parameters for each bit $j$ in parallel.

In the second phase, we sample each $Y_{ij}$ bit-by-bit using \eqref{eq:pS_Y} via
\begin{equation}
\label{eq:Yij_update}
\begin{split}
p(Y_{ij}\:\!\vert\:\!\mathbf{Y}_{-ij},\mathcal{D}) &\propto 
    p(Y_{ij}\:\!\vert\:\!\mathbf{Y}_{-ij})\:\!p(\bar{\mathcal{S}}|\mathbf{Y})
    \approx q^{\backslash ij}(Y_{ij})\:\!p(\bar{\mathcal{S}}|\mathbf{Y})
    = \Phi(\gamma_{ij}Y_{ij})p(\bar{\mathcal{S}}|\mathbf{Y})
\end{split}
\end{equation}
where $\mathbf{Y}_{-ij}$ denotes entries of $\mathbf{Y}$ other than $Y_{ij}$ and $q^{\backslash ij}(Y_{ij})$ denotes a tilted distribution for approximating the posterior from the FITC model; and the probability $q^{\backslash ij}(Y_{ij})$ has the form $\Phi(\gamma_{ij}Y_{ij})$ where $\gamma_{ij}$ is a site parameter computed by the scalable EP algorithm (See \cite{Hernandez:2016} for details). The likelihood $p(\bar{\mathcal{S}}|\mathbf{Y})$ is given in \eqref{eq:pS_Y} for the full version $\mathcal{S}$.

\begin{figure*}[!ht]
    \centering
    \begin{subfigure}[b]{0.23\textwidth}
        \includegraphics[width=\textwidth]{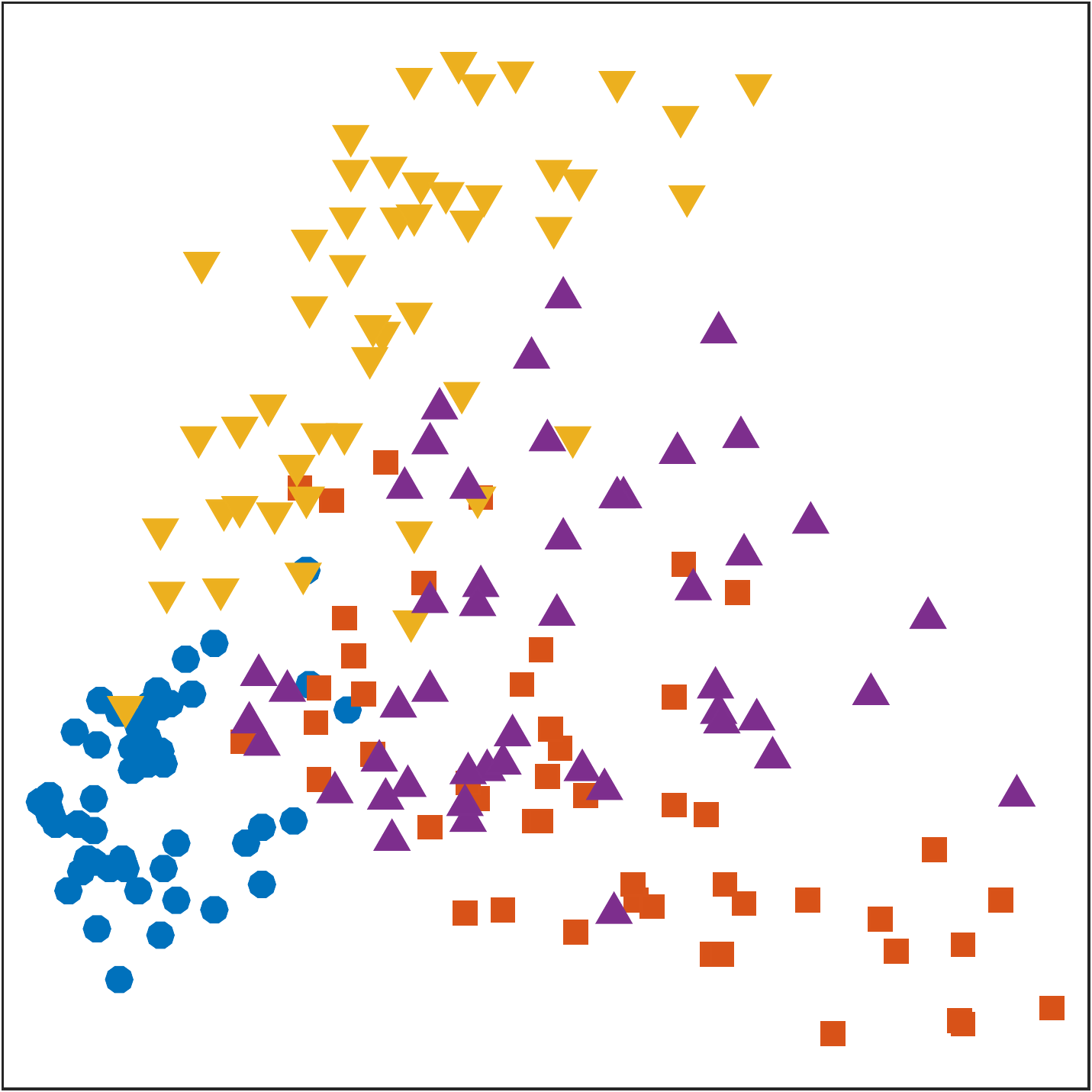}
    \end{subfigure}%
    \quad
    \begin{subfigure}[b]{0.23\textwidth}
        \includegraphics[width=\textwidth]{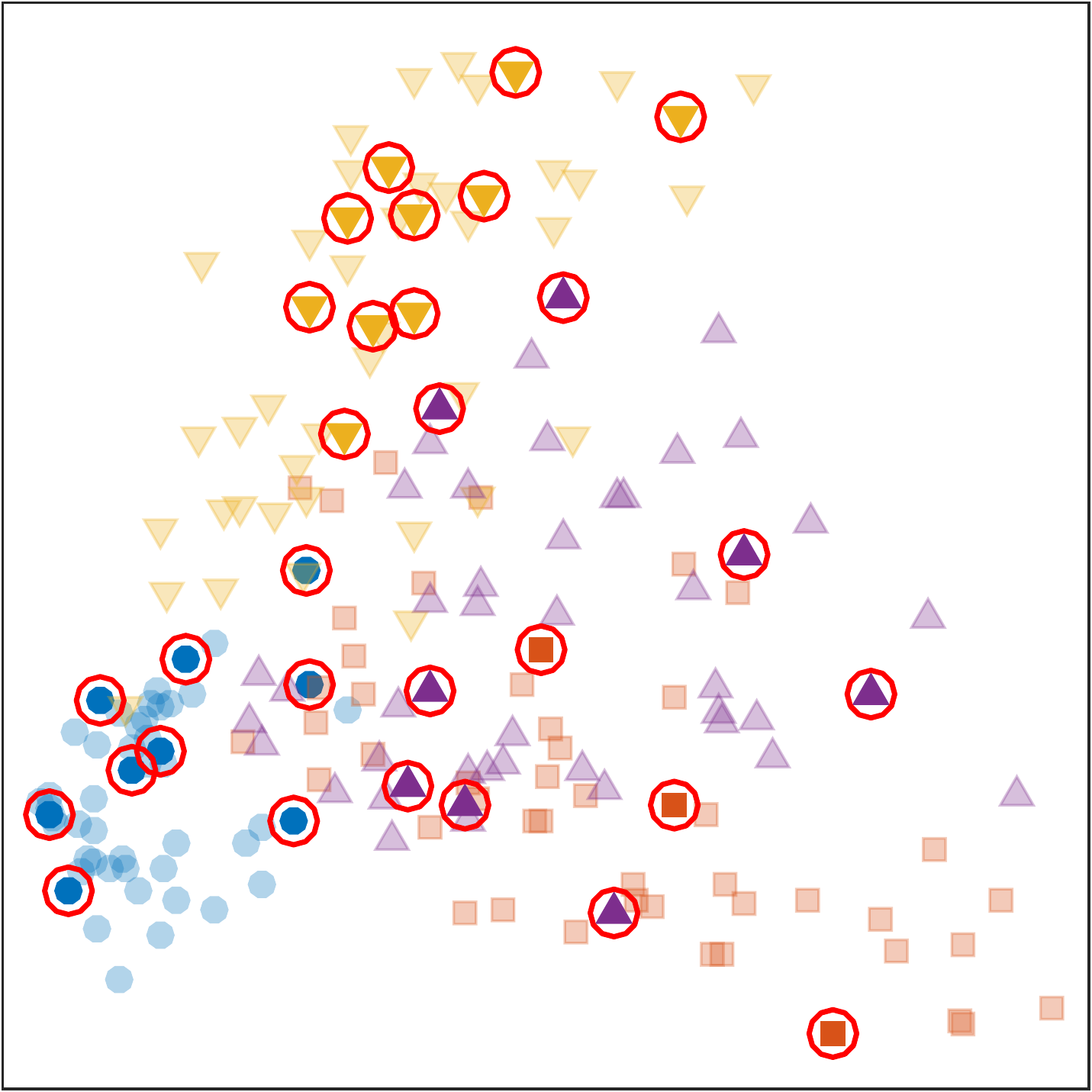}
    \end{subfigure}%
    \quad
    \begin{subfigure}[b]{0.23\textwidth}
        \includegraphics[width=\textwidth]{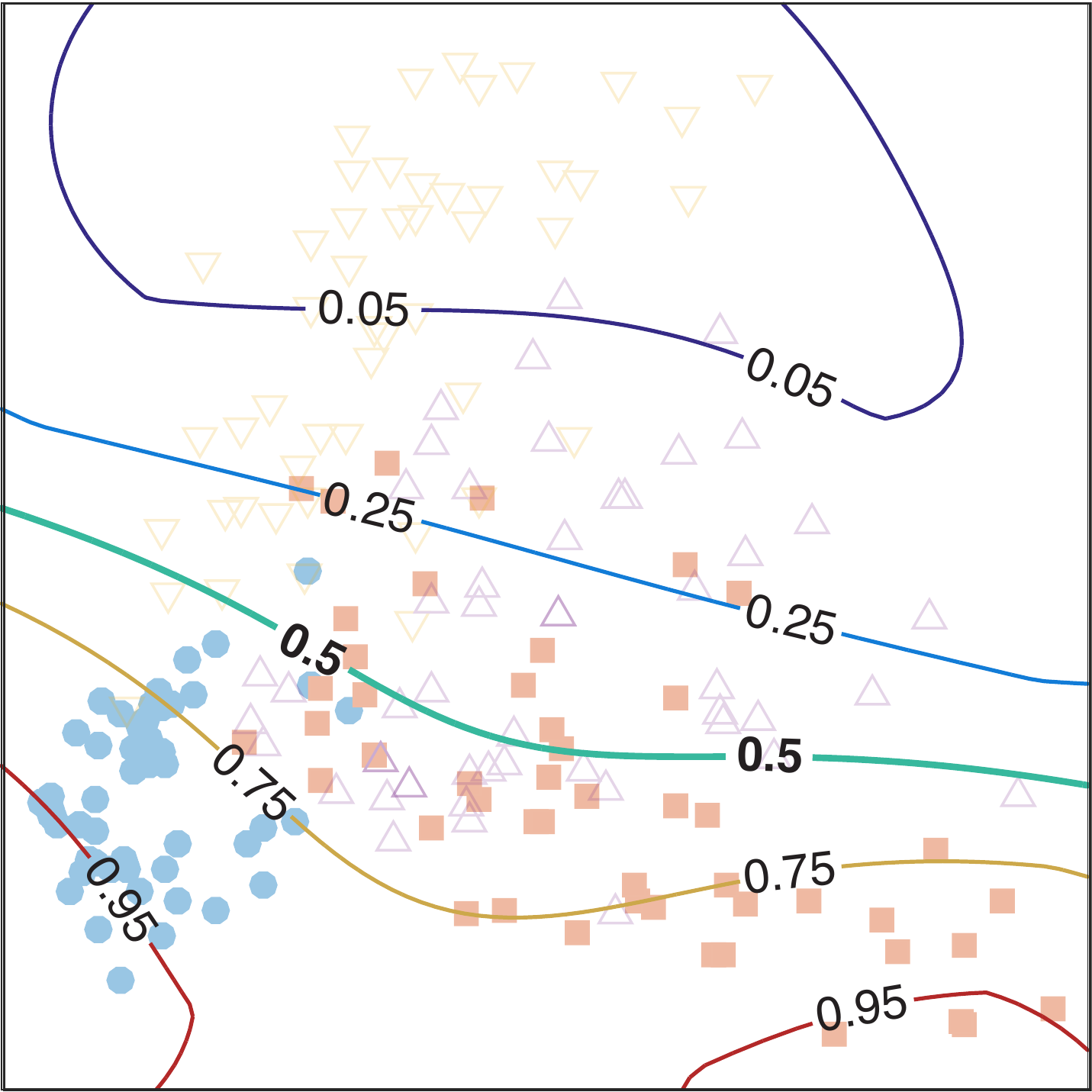}
    \end{subfigure}%
    \quad
    \begin{subfigure}[b]{0.23\textwidth}
        \includegraphics[width=\textwidth]{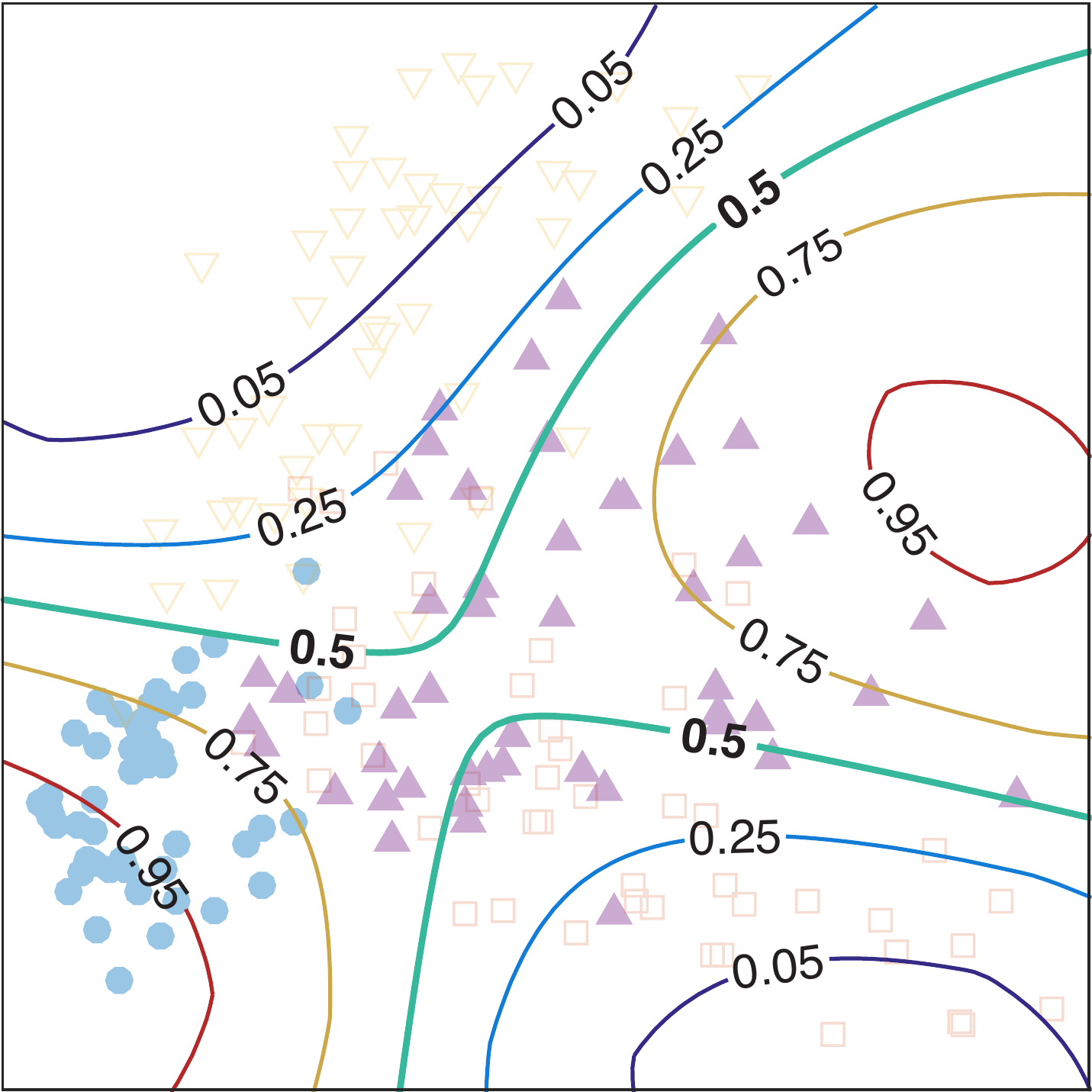}
    \end{subfigure}
    \caption{From left to right; a sample dataset of 4 classes indicated by color and shape $\mathcal{X}$, inducing points marked with red circles $\bar{\mathcal{X}}$, the predictive probability distribution for the first binary encoding $p(y_{*1}|\mathcal{D},\mathbf{x}_*)$ and for the second encoding $p(y_{*2}|\mathcal{D},\mathbf{x}_*)$. The binary codes  $\mathbf{Y}$ by the Gibbs sampler are indicated by filled and empty shapes for $+1$ and $-1$, respectively.}
    \label{fig:example}
\end{figure*}

Our hashing method is illustrated in Figure~\ref{fig:example} for $m=2$ on a two-dimensional sample dataset of 200 points with 4 equal-sized classes and 30 inducing points randomly selected from the dataset.

\section{Experiments}
\label{sec:experiments}

We compared the retrieval performance of our hashing method with several state-of-the-art supervised hashing techniques including CCA-ITQ \cite{Gong:2013kp}, KSH \cite{Liu:2012ii}, FastHash \cite{Lin:2015gs} and SDH \cite{Shen:2015fs}. The public codes provided by the authors are used with their suggested parameters unless otherwise specified. The experiments were performed on three image datasets, namely, CIFAR-10, MNIST and NUS-WIDE datasets in the MATLAB environment on a machine with a 2.8 GHz Intel Core i7 CPU and 16GB RAM.

\subsection{Datasets and Experimental Setup}

The CIFAR-10 dataset \cite{Krizhevsky:2009ak}, which is a labeled subset of the 80M tiny images dataset, has 60,0000 images from 10 classes. There are 50,000 training and 10,000 test images in the dataset. We used a GIST descriptor \cite{Oliva:2001at} of 512 dimensions to represent each image. The MNIST dataset \cite{Lecun:1998yb} contains 28$\times2$8 pixel images of handwritten digits from `0' to `9'. The dataset hash a training set of 60,000 examples and a test set of 10,000 examples. The third dataset, NUS-WIDE \cite{Chua:2009tj}, includes 269,648 images and associated semantic labels of 81 concepts from Flickr such as dog, flower, street, and dancing. The dataset is split into 161,789 training images and 107,859 test images. As \cite{Shen:2015fs}, we consider two images as semantically similar if there exists at least one common associated concept. We used a 500-dimensional bag-of-words representation where the codebook is generated from SIFT descriptors. All datasets were first centered at zero and then each point vector was normalized to unit length.

For NUS-WIDE, we constructed a reduced test set by sampling 10,000 images associated with any of the most frequent 21 labels for all methods. For the KSH method, we uniformly sampled 5,000 images from the training sets of each dataset because the computational complexity of this approach does not allow it to be trained on the entire datasets as we do with other methods (Table~\ref{table:timing}). Similarly, the tree depth of FastHash was set to 2 due to its longer test time (Table~\ref{table:timing}). FastHash was not trained on that dataset due to its large memory requirements. GPH and KSH were used on pairwise similarities on the NUS-WIDE dataset while SDH and CCA-ITQ were trained on label information on that dataset. The $\ell_2$ loss version of SDH was employed in our experiments since there exist no predefined classes for the NUS-WIDE dataset. We uniformly sampled 1000 inducing points (anchor points) from the training set of each dataset. These points were shared by all hashing methods with kernels (GPH, KSH, and SDH) in training. All these methods used an RBF kernel with a kernel width $\ell$ adjusted specifically for each dataset by cross-validation for each method with a kernel.

Hash functions learned from a subset of a large-scale image dataset can be used for computing the binary codes for the entire dataset. Next, these binary codes can be utilized for efficient large-scale image search in the Hamming space \cite{Gionis:1999wf}. Therefore, generalization ability of supervised hashing techniques is critical for the retrieval performance.

For GPH, we adjusted the hyperparameters $\sigma_f$ and $\sigma_y$ by cross-validation on the training set. However, the performance is usually maximized at $\sigma_y=2/m$. For all datasets, 5,000 representative images were randomly selected for $\bar{\mathcal{S}}$. The scalable EP algorithm was run in a stochastic fashion with a block size of 1,000 images. The binary codes $\mathbf{Y}$ were initialized randomly. The GPH inference algorithm was executed until convergence or at most 50 sweeps. In learning hash functions by the GPH model, we observe that the Gibbs sampler for $\mathbf{Y}$ rapidly converges after assigning discriminative binary codes to all classes. The rest of the inference algorithm focuses on training GPs with the scalable EP algorithm on these \emph{fixed} binary codes. A pattern of binary codes emerges even for most values of $\sigma_y$ although some bit assignments might change during the inference algorithm.

After learning hash functions from training data, we computed binary codes for images in the test sets where each dataset contains 10,000 images. Retrieval performance was evaluated by leave-one-out validation on these images only. Each test image was used once as a query while the remaining test images were turned into a retrieval set. As overfitting is a common issue for supervised hashing, this methodology was employed to assess generalization performance of the hashing methods on images that were not used in training, analogous to new images being added to a database or an enormous dataset from which only a subset of images can be employed for learning hash functions.

\subsection{Experimental Results}

\begin{figure*}[!ht]
\centering
\begin{subfigure}{0.58\textwidth}
\includegraphics[width=\linewidth]{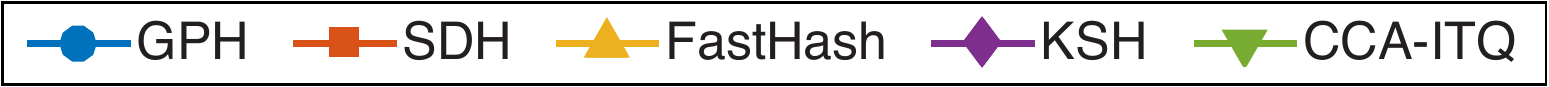}
\end{subfigure}%
\\\vspace{5pt}
\centering
\begin{subfigure}{0.316\textwidth}
\centering
\includegraphics[width=\linewidth]{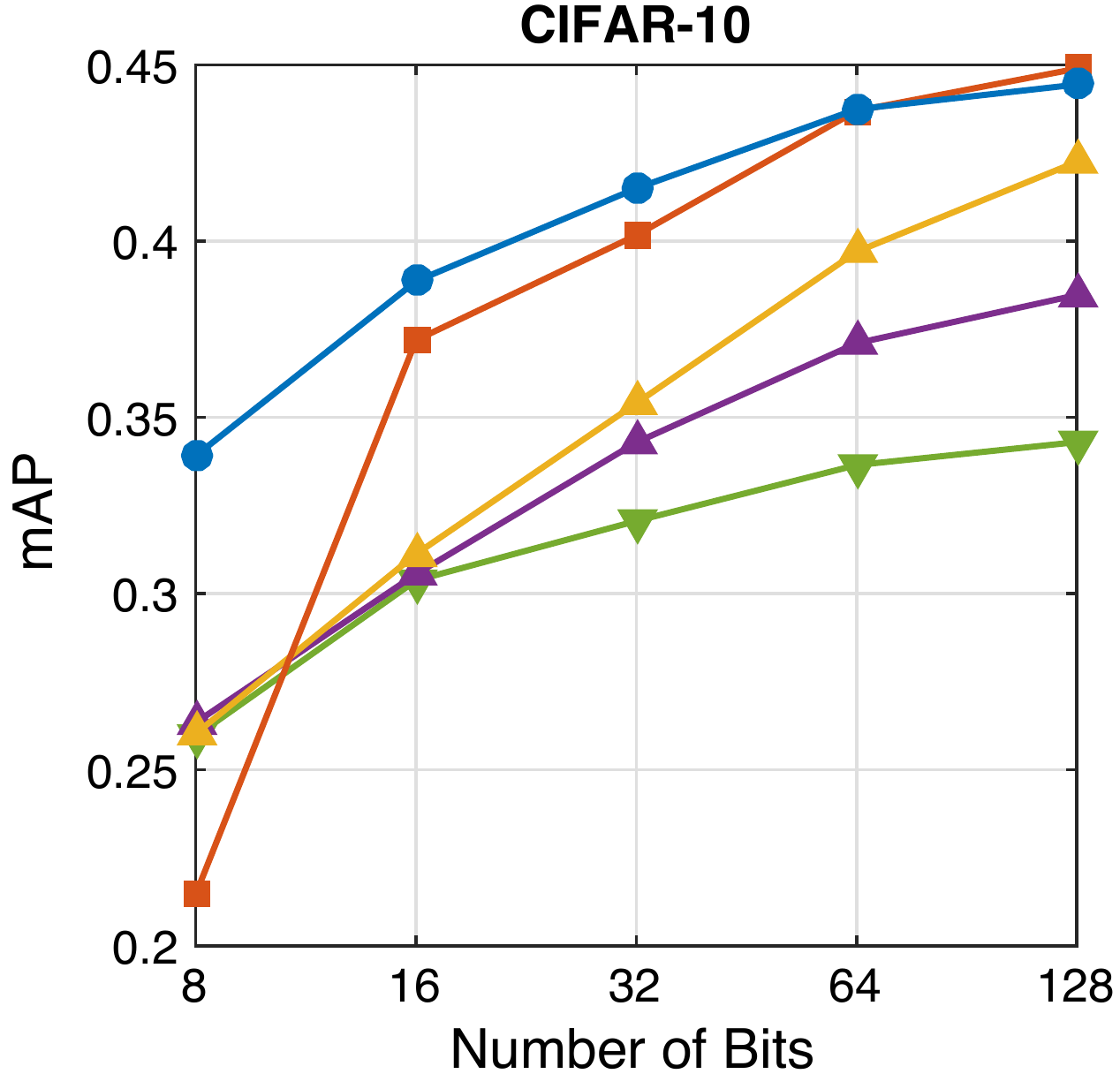}
\end{subfigure}%
\quad
\begin{subfigure}{0.316\textwidth}
\centering
\includegraphics[width=\linewidth]{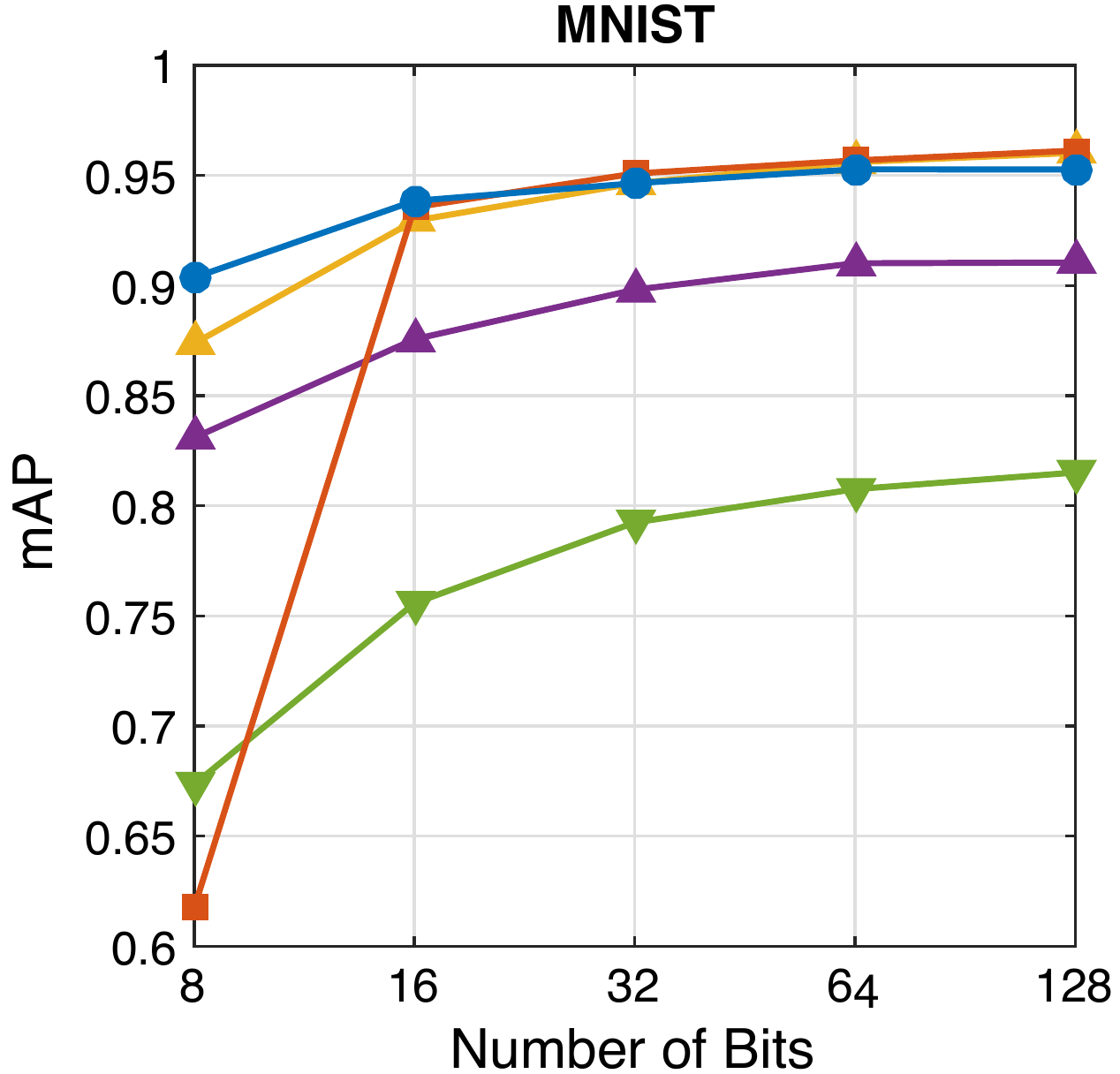}
\end{subfigure}%
\quad
\begin{subfigure}{0.316\textwidth}
\includegraphics[width=\linewidth]{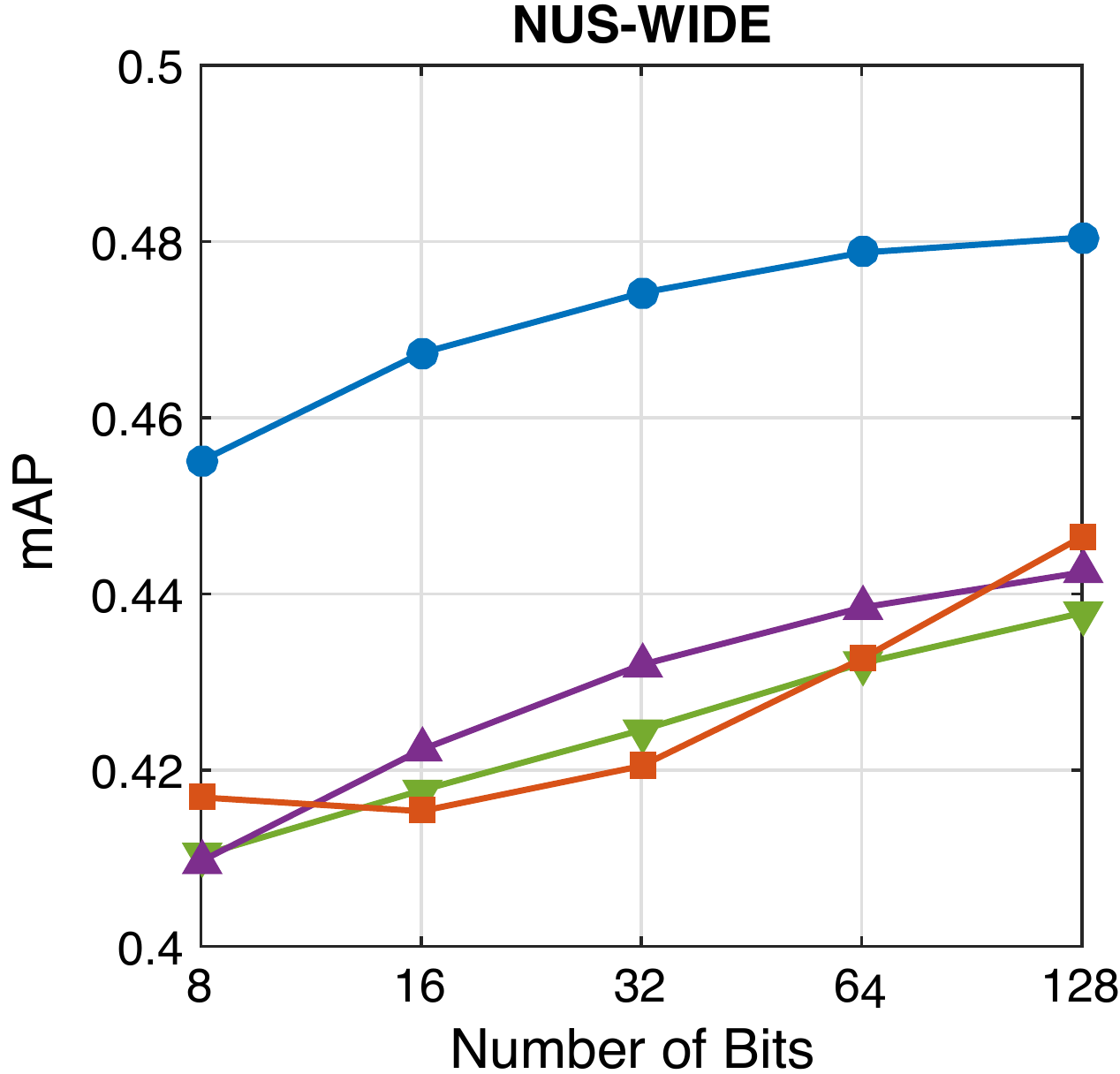}
\end{subfigure}%
\caption{Our method (GPH) is compared with the state-of-the-art methods on, from left to right, the CIFAR-10, MNIST and NUS-WIDE datasets in terms of mean average precision (mAP), respectively.}
\label{fig:retrieval}
\end{figure*}

\begin{figure*}[!ht]
\centering
\begin{subfigure}{0.55\textwidth}
\includegraphics[width=\linewidth]{figures/retrieval_legend}
\end{subfigure}%
\\\vspace{5pt}
\centering
\begin{subfigure}{0.36\textwidth}
\centering
\includegraphics[width=\linewidth]{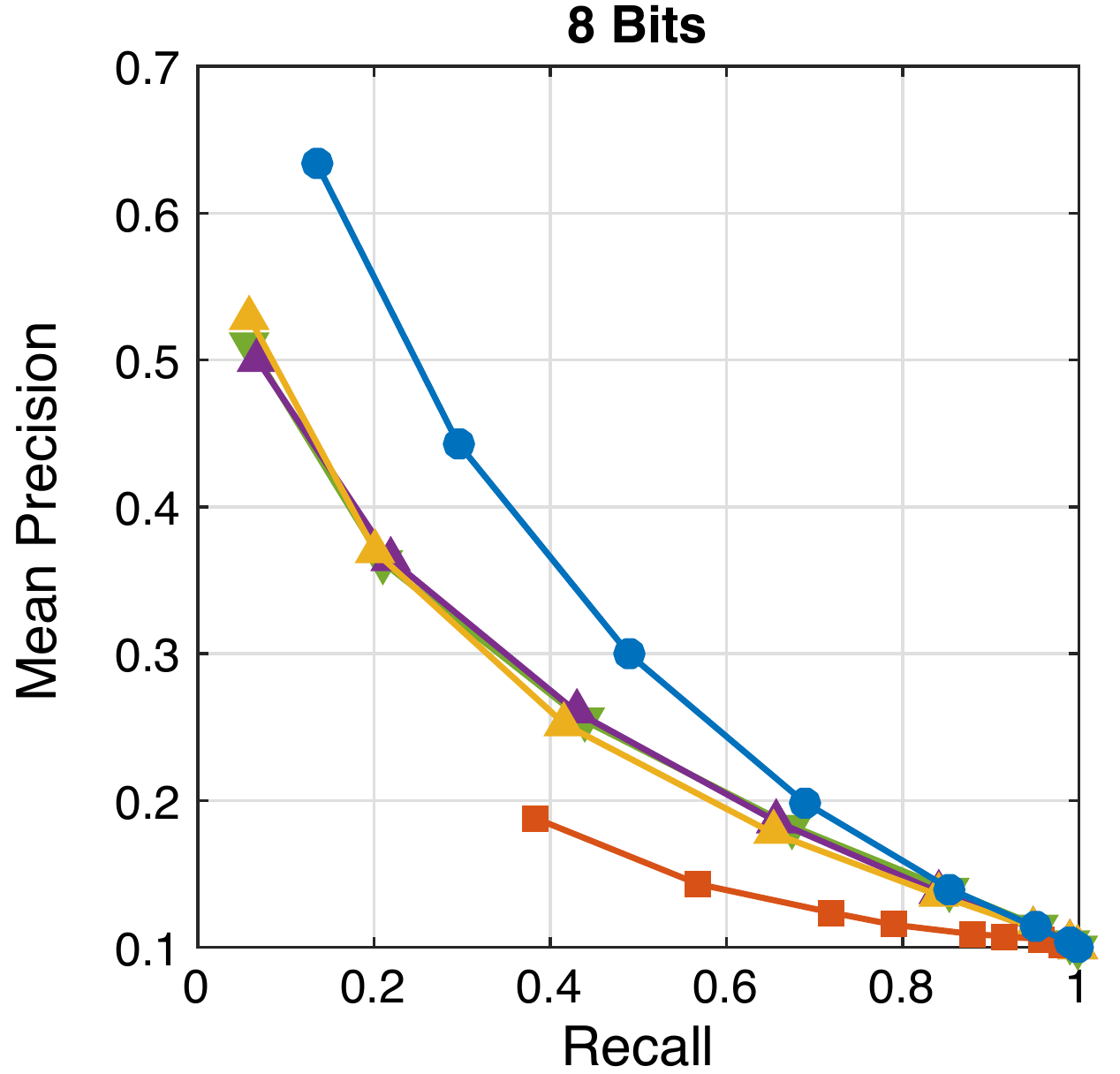}
\end{subfigure}%
\qquad\quad
\begin{subfigure}{0.36\textwidth}
\centering
\includegraphics[width=\linewidth]{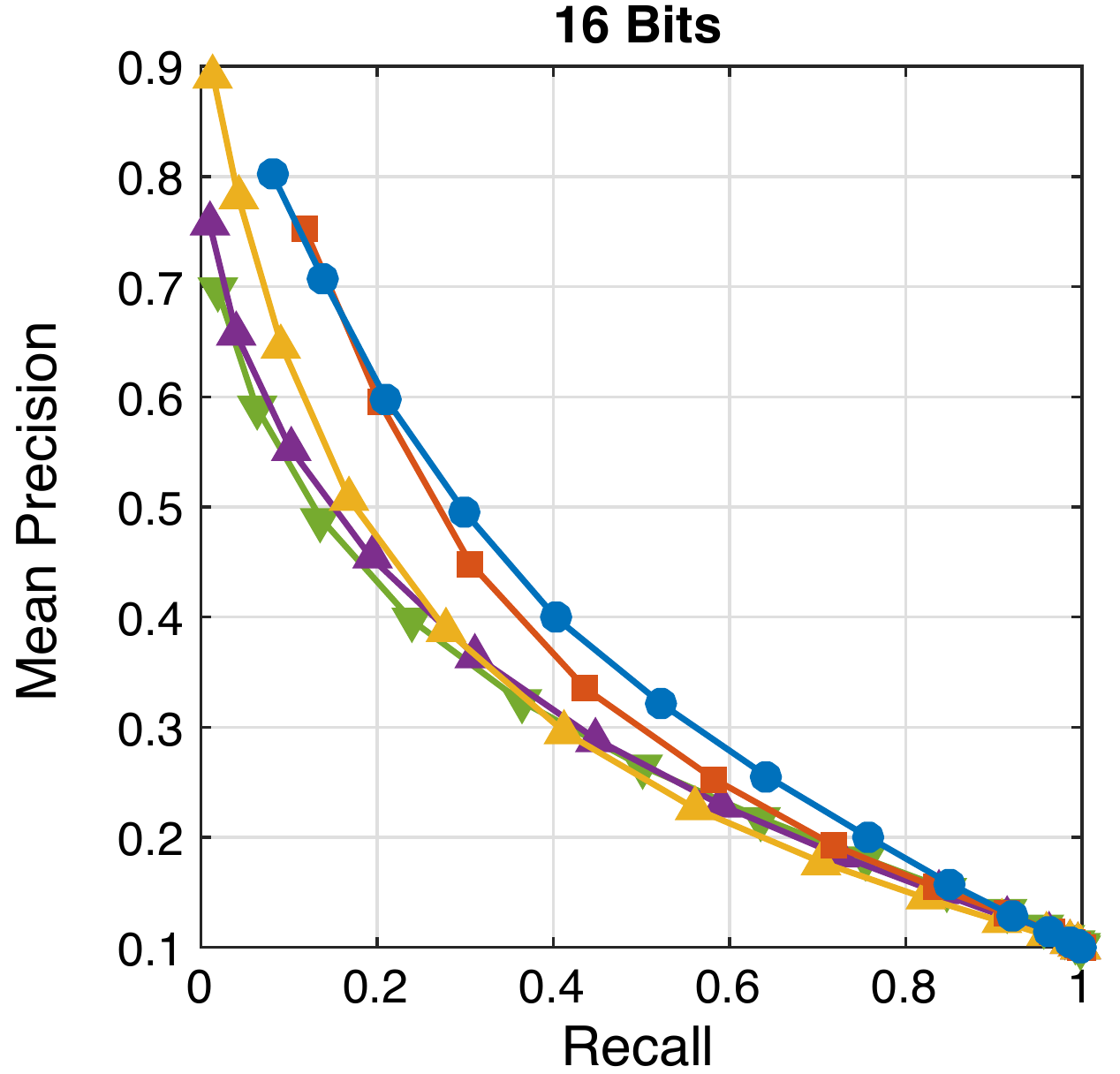}
\end{subfigure}\\
\vspace{12pt}
\begin{subfigure}{0.36\textwidth}
\includegraphics[width=\linewidth]{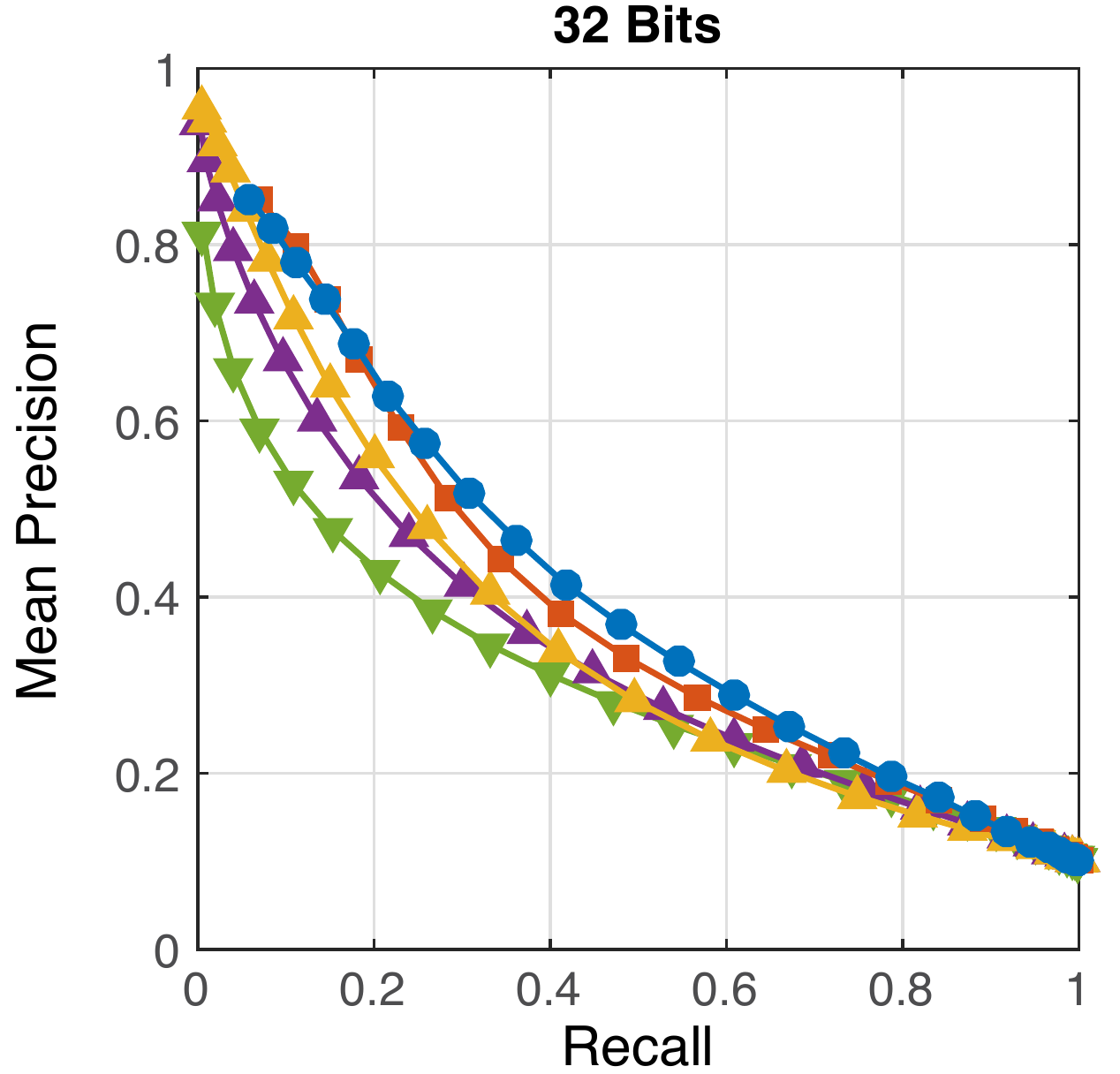}
\end{subfigure}%
\qquad\quad
\begin{subfigure}{0.36\textwidth}
\includegraphics[width=\linewidth]{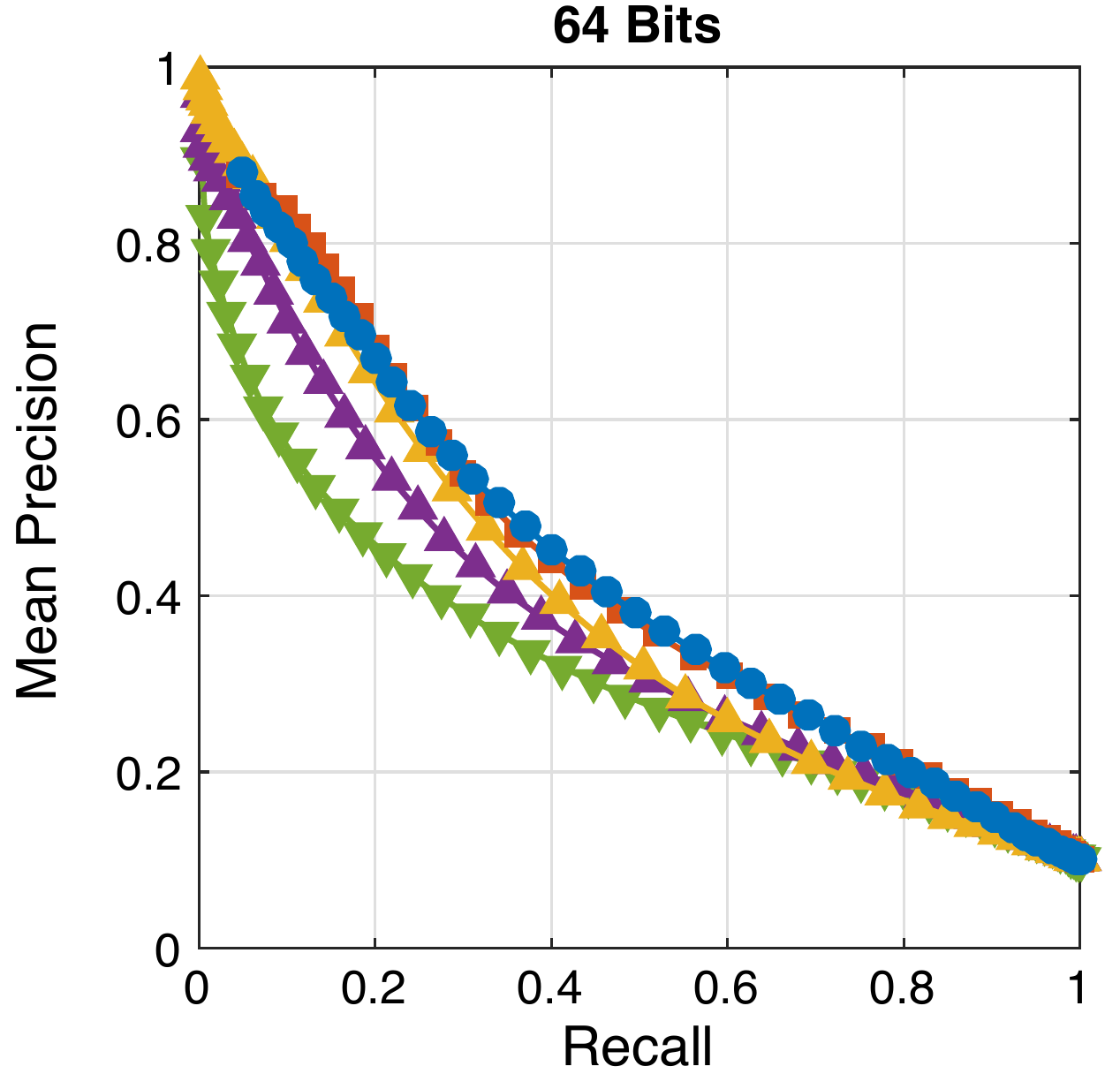}
\end{subfigure}%
\caption{Our method (GPH) is compared with the state-of-the-art methods on the CIFAR-10 datasets by precision-recall curves for 8, 16, 32 and 64 bit hash codes.}
\label{fig:recprec}
\end{figure*}

For each query, images were ranked concerning Hamming distances between their binary codes and that of the given query. Ground truth labels were defined by semantic similarity from either class labels or associated tag information. For quantitative analysis, we report retrieval performance in mean average precision (mAP) for all datasets in Figure~\ref{fig:retrieval}. Precision-recall curves of all methods on the CIFAR-10 datasets are demonstrated in Figure~\ref{fig:recprec} for different lengths of hash codes.

Our method GPH outperforms the state-of-the-art supervised hashing methods in terms of mAP for smaller binary codes (32-bits or shorter) on all datasets. Some methods perform better than GPH in some cases on the training set (data not shown); however, GPH outperforms on the independent set due to better generalization with the help of averaging over nonlinear classifiers. Although SDH performs well on the CIFAR-10 and MNIST datasets, its performance on the NUS-WIDE dataset where there are no predefined classes is poor. On the other hand, GPH performs the best on that dataset with a large margin. As the number of bits increases, the approximation error due to our assumption of conditional independence between GPs is growing as well. Therefore, the improvement in retrieval performance by longer codes for the GPH is moderate.

The comparative retrieval performance of our method is shown in Table~\ref{table:timing} with different training and inducing set sizes along with the state-of-the-art methods on the CIFAR-10 dataset for 16-bit hash codes in terms of mAP and mean precision at Hamming radius 2. The execution times are reported for training and computing binary code of a single query using the learned hashing functions. Larger training and inducing sets provide better performance with longer execution time. The number of inducing points affects retrieval performance than training size. Note that the scale of the inducing set also affects the test time. Our method provides better performance in retrieval. Note that Table~\ref{table:timing} shows the training time on a four-core machine and our method learns its hash function by a parallel algorithm. The training time in Table~\ref{table:timing} can be improved by a computer with a larger number of nodes. Therefore, GPH is more efficient and effective in retrieval when binary codes are short. CCA-ITQ is the fastest method in the experiments; however, its performance is the worst in retrieval. On the other hand, SDH is the second most efficient method with relatively good retrieval performance. For testing, all kernel methods (GPH, SDH, and KSH) have a similar number of operations: Kernel vector computation, multiplication by weight matrix and kernel vector and finally sign operation. Therefore, they have similar test time per query. CCA-ITQ, which has no kernel computation, has faster test time. On the other hand, FastHash has longer test time due to its computations based on decision trees.

\begin{table*}[!ht]
  \caption{Results in mean average precision (mAP), mean of precision at Hamming radius $r = 2$, training and test time for 16-bit hash codes on the CIFAR-10 dataset. The experiments were performed on a machine with an Intel quad-core processor.} \label{table:timing}
\vskip 0.05in
\begin{center}
  \begin{tabular}{|l||r|r|r|r|r|r|}
  \hline
   \multicolumn{1}{|c||}{\multirow{2}{*}{\bfseries Method}} & \multicolumn{1}{c|}{\bfseries Training} & \multicolumn{1}{c|}{\bfseries Inducing} &  \multicolumn{1}{c|}{\multirow{2}{*}{\bfseries mAP}} & \multicolumn{1}{c|}{\bfseries Precision} & \multicolumn{1}{c|}{\bfseries Training} & \multicolumn{1}{c|}{\bfseries Test Time}\\
   & \multicolumn{1}{c|}{\bfseries Set Size} & \multicolumn{1}{c|}{\bfseries Set Size} &  & \multicolumn{1}{c|}{\bfseries at $r=2$} & \multicolumn{1}{c|}{{\bfseries Time} (sec)} & \multicolumn{1}{c|}{($\mu$-sec$/$query)}\\
    \hline
    \hline
    \multicolumn{1}{|l||}{\multirow{6}{*}{GPH}}& 5,000 & 300 & 0.279 & 0.365 & 26.5 & 7.7\\
    & 5,000 & 500 & 0.304 & 0.387 & 37.6 & 12.3 \\
    & 5,000 & 1,000 & 0.309 & 0.395 & 78.5 & 21.6 \\
    & 5,000 & 3,000 & 0.338 & 0.416 & 456.1 & 44.7\\
    & 50,000 & 300 & 0.311 & 0.403 & 177.4 & 7.6 \\
    & 50,000 & 1,000 & \textbf{0.395} & \textbf{0.466} & 497.9 & 18.9 \\
    \hline
    \multicolumn{1}{|l||}{\multirow{2}{*}{SDH}} & 5,000 & 1,000 & 0.306 & 0.373 & 0.9 & 17.1 \\
    & 50,000 & 1,000 & 0.372 & 0.423 & 7.5 & 16.5 \\
    \hline 
    \multicolumn{1}{|l||}{\multirow{2}{*}{Fasthash}} & 5,000 & \multicolumn{1}{|c|}{\quad-} & 0.240 & 0.357 & 24.6 & 47.5 \\
    & 50,000 & \multicolumn{1}{|c|}{\quad-} & 0.311 & 0.440 & 355.5 & 68.0 \\
    \hline    
    KSH & 5,000 & 1,000 & 0.305 & 0.408 & 1,461.6 & 25.0\\   
    \hline     
    \multicolumn{1}{|l||}{\multirow{2}{*}{CCA-ITQ}} & 5,000 & \multicolumn{1}{|c|}{\quad-} & 0.262 & 0.353 & 0.1 & 0.2 \\                        
    & 50,000 & \multicolumn{1}{|c|}{\quad-} & 0.304 & 0.389 & 0.3 & 0.2 \\
    \hline
  \end{tabular}
\end{center}
\vskip -0.1in
\end{table*}

\section{Conclusion}
\label{sec:conclusion}

We proposed a supervised retrieval scheme based on Gaussian processes for classification. We developed an efficient inference algorithm for the proposed model. The experimental results on three real-world image datasets show that our method produces the best retrieval performance for smaller binary codes by preventing overfitting to training data.

\subsubsection*{Acknowledgments}

This work was partially supported by the US Government through ONR MURI Grant N000141010934.

\begin{small}

\bibliographystyle{abbrv}
\bibliography{gph_nips16}

\begin{thebibliography}{10}

\bibitem{Candela:2005wp}
J.~Q. Candela and C.~E. Rasmussen.
\newblock {A Unifying View of Sparse Approximate Gaussian Process Regression.}
\newblock {\em Journal of Machine Learning Research (JMLR)}, 6:1939--1959, Dec.
  2005.

\bibitem{Chua:2009tj}
T.-S. Chua, J.~Tang, R.~Hong, H.~Li, Z.~Luo, and Y.-T. Zheng.
\newblock Nus-wide: A real-world web image database from national university of
  singapore.
\newblock In {\em ACM International Conference on Image and Video Retrieval
  (CIVR)}, Santorini, Greece, 2009.

\bibitem{DoshiVelez:2009tf}
F.~Doshi-Velez, D.~A. Knowles, S.~Mohamed, and Z.~Ghahramani.
\newblock {Large Scale Nonparametric Bayesian Inference: Data Parallelisation
  in the Indian Buffet Process}.
\newblock In {\em Advances in Neural Information Processing Systems (NIPS)},
  pages 1294--1302, 2009.

\bibitem{Gelman:2014ww}
A.~Gelman, A.~Vehtari, P.~Jylanki, C.~Robert, N.~Chopin, and J.~P. Cunningham.
\newblock {Expectation propagation as a way of life}.
\newblock {\em arXiv.org}, Dec. 2014.

\bibitem{Gionis:1999wf}
A.~Gionis, P.~Indyk, and R.~Motwani.
\newblock Similarity search in high dimensions via hashing.
\newblock In {\em International Conference on Very Large Data Bases (VLDB)},
  pages 518--529, 1999.

\bibitem{Gong:2013kp}
Y.~Gong, S.~Lazebnik, A.~Gordo, and F.~Perronnin.
\newblock {Iterative quantization: a Procrustean approach to learning binary
  codes for large-scale image retrieval.}
\newblock {\em IEEE Transactions on Pattern Analysis and Machine Intelligence
  (TPAMI)}, 35(12):2916--2929, Dec. 2013.

\bibitem{Hernandez:2016}
D.~Hernandez-Lobato and J.~M. Hernandez-Lobato.
\newblock {Scalable Gaussian Process Classification via Expectation
  Propagation}.
\newblock In {\em International Conference on Artificial Intelligence and
  Statistics (AISTATS)}, 2016.

\bibitem{Krizhevsky:2009ak}
A.~Krizhevsky.
\newblock Learning multiple layers of features from tiny images.
\newblock Technical report, University of Toronto, 2009.

\bibitem{Kulis:2012ey}
B.~Kulis and K.~Grauman.
\newblock {Kernelized Locality-Sensitive Hashing}.
\newblock {\em IEEE Transactions on Pattern Analysis and Machine Intelligence
  (TPAMI)}, 34(6):1092--1104, 2011.

\bibitem{Kuss:2005wx}
M.~Kuss and C.~E. Rasmussen.
\newblock {Assessing Approximate Inference for Binary Gaussian Process
  Classification}.
\newblock {\em Journal of Machine Learning Research (JMLR)}, 6:1679--1704, Oct.
  2005.

\bibitem{Lecun:1998yb}
Y.~Lecun, L.~Bottou, Y.~Bengio, and P.~Haffner.
\newblock Gradient-based learning applied to document recognition.
\newblock {\em Proceedings of the IEEE}, 86(11):2278--2324, Nov 1998.

\bibitem{Lin:2015gs}
G.~Lin, C.~Shen, Q.~Shi, A.~Van~den Hengel, and D.~Suter.
\newblock Fast supervised hashing with decision trees for high-dimensional
  data.
\newblock In {\em IEEE Conference on Computer Vision and Pattern Recognition
  (CVPR)}, pages 1971--1978, June 2014.

\bibitem{Liu:2012ii}
W.~Liu, J.~Wang, R.~Ji, Y.-G. Jiang, and S.-F. Chang.
\newblock {Supervised hashing with kernels.}
\newblock In {\em IEEE Conference on Computer Vision and Pattern Recognition
  (CVPR)}, pages 2074--2081, 2012.

\bibitem{Minka:2001we}
T.~P. Minka.
\newblock {Expectation Propagation for approximate Bayesian inference}.
\newblock In {\em Conference in Uncertainty in Artificial Intelligence (UAI)},
  Aug. 2001.

\bibitem{NaishGuzman:2007uf}
A.~Naish-Guzman and S.~Holden.
\newblock {The generalized FITC approximation}.
\newblock In {\em Advances in Neural Information Processing Systems (NIPS)},
  pages 1057--1064, 2007.

\bibitem{Nickisch:2008vx}
H.~Nickisch and C.~E. Rasmussen.
\newblock {Approximations for binary Gaussian process classification}.
\newblock {\em Journal of Machine Learning Research (JMLR)}, 9:2035--2078, Oct.
  2008.

\bibitem{Norouzi:2011mf}
M.~Norouzi and D.~J. Fleet.
\newblock Minimal loss hashing for compact binary codes.
\newblock In {\em International Conference on Machine Learning (ICML)}, 2011.

\bibitem{Oliva:2001at}
A.~Oliva and A.~Torralba.
\newblock Modeling the shape of the scene: A holistic representation of the
  spatial envelope.
\newblock {\em International Journal of Computer Vision (IJCV)},
  42(3):145--175, 2001.

\bibitem{Ozdemir:2014wi}
B.~Ozdemir and L.~S. Davis.
\newblock {A Probabilistic Framework for Multimodal Retrieval using Integrative
  Indian Buffet Process}.
\newblock In {\em Advances in Neural Information Processing Systems (NIPS)},
  pages 2384--2392, 2014.

\bibitem{Rasmussen:2006vz}
C.~E. Rasmussen and C.~K.~I. Williams.
\newblock {\em {Gaussian processes for machine learning}}.
\newblock MIT Press, Cambridge, Massachusetts, 2006.

\bibitem{Rastegari:2013ue}
M.~Rastegari, J.~Choi, S.~Fakhraei, H.~D. III, and L.~S. Davis.
\newblock {Predictable Dual-View Hashing}.
\newblock {\em International Conference on Machine Learning (ICML)}, pages
  1328--1336, 2013.

\bibitem{Shen:2015fs}
F.~Shen, C.~Shen, W.~Liu, and H.~Tao~Shen.
\newblock Supervised discrete hashing.
\newblock In {\em IEEE Conference on Computer Vision and Pattern Recognition
  (CVPR)}, June 2015.

\bibitem{Snelson:2005wc}
E.~Snelson and Z.~Ghahramani.
\newblock {Sparse Gaussian Processes using Pseudo-inputs}.
\newblock In {\em Advances in Neural Information Processing Systems (NIPS)},
  pages 1257--1264, 2005.

\bibitem{Nitish:2012dl}
N.~Srivastava and R.~Salakhutdinov.
\newblock Multimodal learning with deep boltzmann machines.
\newblock In {\em Advances in Neural Information Processing Systems (NIPS)},
  pages 2222--2230, 2012.

\bibitem{Weiss:2008tu}
A.~Torralba, R.~Fergus, and Y.~Weiss.
\newblock Small codes and large image databases for recognition.
\newblock In {\em IEEE Conference on Computer Vision and Pattern Recognition
  (CVPR)}, pages 1--8, June 2008.

\bibitem{Wang:2010jk}
J.~Wang, S.~Kumar, and S.-F. Chang.
\newblock Sequential projection learning for hashing with compact codes.
\newblock In {\em International Conference on Machine Learning (ICML)}, pages
  1127--1134, 2010.

\bibitem{Yildirim2012-YILARA}
I.~Yildirim and R.~A. Jacobs.
\newblock A rational analysis of the acquisition of multisensory
  representations.
\newblock {\em Cognitive Science}, 36(2):305--332, 2012.

\bibitem{Zhang:2014jr}
P.~Zhang, W.~Zhang, W.-J. Li, and M.~Guo.
\newblock {Supervised hashing with latent factor models}.
\newblock {\em International ACM SIGIR Conference on Research and Development
  in Information Retrieval}, pages 173--182, 2014.

\end{thebibliography}

\end{small}

\end{document}